\documentclass[a4paper, 10pt, conference]{ieeeconf}      
\usepackage{FG2026}
\usepackage{amsmath}
\usepackage{graphicx}
\usepackage{subcaption}
\usepackage{tablefootnote}
\usepackage{multirow}
\usepackage{booktabs}
\usepackage{hyperref}

\FGfinalcopy 

\IEEEoverridecommandlockouts                              
\def\FGPaperID{****} 

\title{\LARGE \bf
Arc2Morph: Identity-Preserving Facial Morphing with Arc2Face
}

\author{\parbox{16cm}{\centering
    {\large Nicolò Di Domenico, Annalisa Franco, Matteo Ferrara, Davide Maltoni}\\
    {\normalsize
    Department of Computer Science and Engineering, University of Bologna, Italy}}
    \thanks{This project received funding from the European Union’s Horizon Europe research and innovation program under Grant Agreement No.101121280. This text reflects only the author’s views, and the commission is not liable for any use that may be made of the information contained therein.}
}

\begin{document}

\ifFGfinal
\thispagestyle{empty}
\pagestyle{empty}
\else
\author{Anonymous FG2026 submission\\ Paper ID \FGPaperID \\}
\pagestyle{plain}
\fi
\maketitle

\begin{abstract}

Face morphing attacks are widely recognized as one of the most challenging threats to face recognition systems used in electronic identity documents. These attacks exploit a critical vulnerability in passport enrollment procedures adopted by many countries, where the facial image is often acquired without a supervised live capture process. In this paper, we propose a novel face morphing technique based on Arc2Face, an identity-conditioned face foundation model capable of synthesizing photorealistic facial images from compact identity representations. We demonstrate the effectiveness of the proposed approach by comparing the morphing attack potential metric on two large-scale sequestered face morphing attack detection datasets against several state-of-the-art morphing methods, as well as on two novel morphed face datasets derived from FEI and ONOT. Experimental results show that the proposed deep learning–based approach achieves a morphing attack potential comparable to that of landmark-based techniques, which have traditionally been regarded as the most challenging. These findings confirm the ability of the proposed method to effectively preserve and manage identity information during the morph generation process.

\end{abstract}

\section{INTRODUCTION}
Face morphing attacks are widely recognized as one of the most challenging threats to Face Recognition Systems (FRSs) used in electronic identity documents. These attacks exploit a critical weakness in the enrollment procedures adopted by many countries, where the facial image is often acquired without a supervised live capture process. Under these conditions, two individuals may collaborate to create a single morphed facial image that combines identity features from both subjects. This image can be presented during the document issuance phase to deceive the human officer responsible for identity verification, resulting in the storage of a double-identity image within the document’s chip.
The severity of this attack lies in the intrinsic vulnerability of face verification systems: due to the blended nature of the morphed image, it can successfully match against both contributing subjects. Consequently, if the morphing attack is not detected during enrollment, two different individuals may later authenticate as the same identity holder, effectively sharing a single regular identity document. 
Two conditions must be met for an attack to be successful: (i) the morphed image must convincingly deceive a human examiner by showing high similarity to the document applicant and maintaining strong visual realism, with no noticeable artifacts or defects; (ii) at the same time, the image must also deceive the FRS used for automatic identity verification, allowing it to be successfully matched to both individuals.

Since the introduction of face morphing attacks in 2014 \cite{magicPassport}, significant research efforts have focused on developing effective detection methods. These approaches operate either on a single image, suitable for the enrollment stage, or on an image pair, which is more feasible during identity verification at the borders. In parallel with advancements in detection techniques, face morphing generation methods have also evolved continuously, becoming increasingly sophisticated and harder to detect.
Existing morphing techniques can be broadly classified into two main categories \cite{Ferrara2022}. The first category comprises landmark-based approaches, where the morphing process relies on extracting corresponding facial landmarks that are used to geometrically warp the two contributing faces, followed by a texture blending step. The second category includes deep learning–based techniques, which exploit the generative capabilities of neural networks to synthesize a morphed facial image directly from the face images of two different subjects.
For a long time, landmark-based approaches were considered the most effective solution, as they can generate high-quality morphed images while reliably preserving the identity information from both contributing subjects. More recently, however, the rapid advancement of deep generative models has significantly increased interest in deep learning–based morphing techniques, which are now capable of producing highly realistic and identity-preserving morphs. 

\begin{figure}
    \centering
    \includegraphics[width=0.32\linewidth]{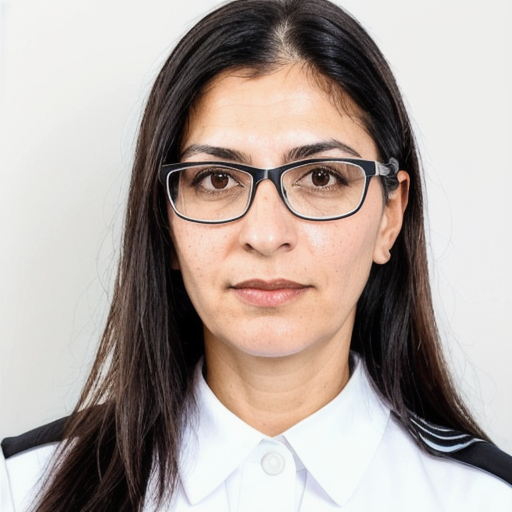}
    \hfill
    \includegraphics[width=0.32\linewidth]{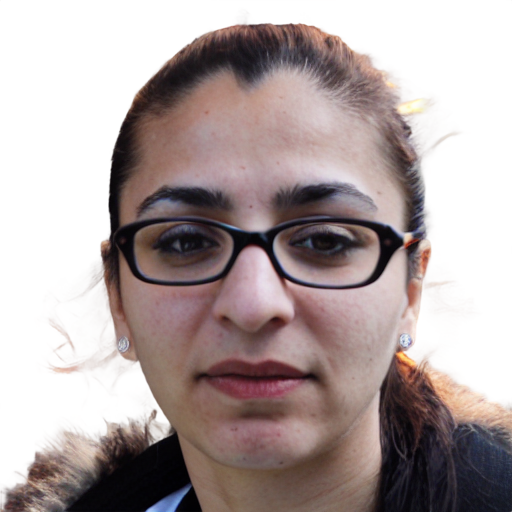}
    \hfill
    \includegraphics[width=0.32\linewidth]{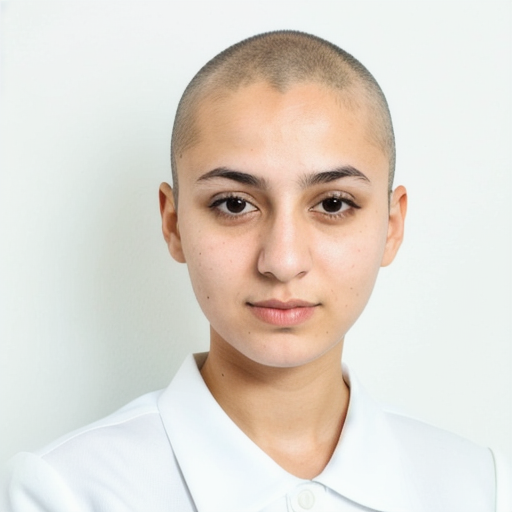} \\
    \vspace{2mm}
    \includegraphics[width=0.32\linewidth]{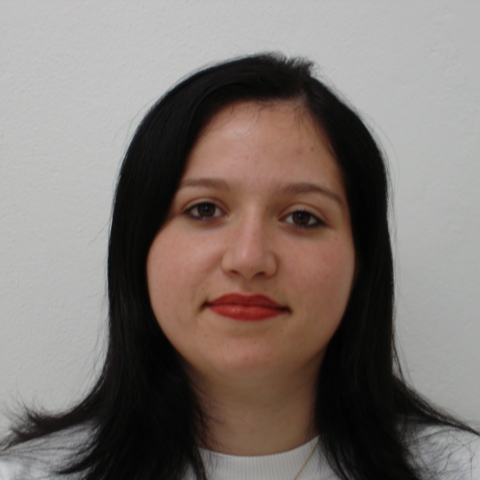}
    \hfill
    \includegraphics[width=0.32\linewidth]{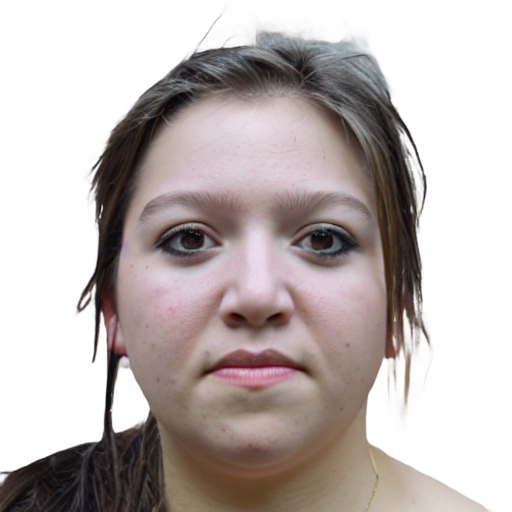}
    \hfill
    \includegraphics[width=0.32\linewidth]{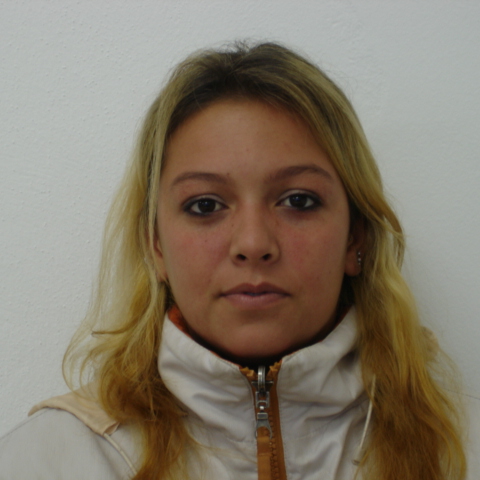}
    \caption{Samples of morphed images (center column) generated with our proposed method. On the first row, Accomplice (left column) and criminal (right column) are sourced from the ONOT~\cite{di2024onot} dataset; on the second row, accomplice and criminal are sourced from the FEI~\cite{Fei_THOMAZ2010902} dataset.}
    \label{fig:eye}
\end{figure}

Since identity preservation represents a critical challenge for deep learning–based morphing approaches, this paper investigates the feasibility of leveraging a model specifically designed to capture representative identity features, namely Arc2Face, for face morphing generation (see Figure \ref{fig:eye} for some examples). Particular attention is given to the generation of ISO/ICAO compliant images, feasible for a realistic enrollment process, by explicitly controlling some characteristics of the generated image such as the pose and the background.
In particular, the main contributions of this work are as follows:
\begin{itemize}
    \item a novel deep learning-based morphing approach that achieves attack potential comparable to, and in some cases exceeding, that of landmark-based techniques, while clearly outperforming existing State-Of-The-Art (SOTA) deep learning-based morphing methods;
    \item an extensive evaluation and comparison of several SOTA morphing approaches on both real and synthetic face images, providing an up-to-date overview of the current landscape of morphing attacks;
    \item the public release of the implementation to ensure full reproducibility and to foster further research on this topic;
    \item two newly generated datasets of morphed facial images, created using the proposed method and made publicly available to the research community for benchmarking and research purposes.
\end{itemize}

The following sections provide a brief review of the SOTA in face morphing, followed by a description of the proposed approach and the experimental evaluation carried out to evaluate its effectiveness. 

\section{RELATED WORKS}
The literature includes a wide range of face morphing approaches which can be classified into two main categories: landmark-based or deep learning-based \cite{Ferrara2022,FrsMaSur,venkatesh2021face} methods.

\subsection{Landmark-based approaches}
Landmark-based face morphing approaches synthesize smooth and gradual transformations between two facial images by relying on facial landmarks extracted from the input images. These landmarks are typically detected using publicly available models, such as DLib \cite{dlib09}, and correspond to salient facial features (e.g., eyes, nose, mouth, and eyebrows) and are used to approximate the shape and structure of the face. The morphing process is generally realized through a combination of geometric warping and texture blending. Specifically, image warping is applied to align the facial geometry of the two subjects according to the extracted landmarks, while texture blending merges their visual appearance to produce a coherent morphed image \cite{8897253,demorphing}. One of the main advantages of these methods is the high degree of control they offer over the resulting identity, as the transformation can be explicitly guided by the precise positioning of facial features.
However, the quality of landmark-based morphed images strongly depends on the accuracy of landmark detection. In case where landmarks are imprecise or absent, visual artifacts often appear, particularly around critical facial areas such as the eyes, nose, and mouth. To address these limitations, recent research efforts have increasingly focused on the development of post-processing methods aimed at automatically removing visible artifacts \cite{9558766, 10593948}, as well as on improved morphing approaches able to produce higher-quality and artifact-free images \cite{makrushin2017automatic,10593985}.

\subsection{Deep learning-based approaches}
Deep-learning-based face morphing approaches aim to generate morphed images while avoiding the artifacts typically associated with landmark-based methods. These techniques exploit the generative capabilities of models such as Generative Adversarial Networks (GANs) and, more recently, diffusion models, to synthesize morphed faces by jointly sampling from the latent representations of the two contributing facial images. Generally, these methods achieve high visual quality; however, they often provide less explicit control over the resulting identity and may still introduce artifacts, typical for instance of GAN-based generation.

MorGAN \cite{8698563} represents one of the earliest GAN-based face morphing methods and is inspired by the Bidirectional Generative Adversarial Network (BiGAN) architecture \cite{donahue2017adversarialfeaturelearning}, which introduces an encoder to map images into a latent space. By jointly discriminating image–latent pairs, the BiGAN framework encourages the encoder to invert the generator, MorGAN adapts this idea to enable face morph generation. Despite its conceptual relevance, MorGAN is constrained by the low resolution of the generated images ($64\times64$ pixels), which is insufficient to satisfy ISO/ICAO quality standards~\cite{icao-portrait,ISO-19794-5,ISO-29794-5,ISO-39794-5}.
To overcome this limitation, subsequent work focused on the generation of high-resolution morphed images (up to $1024\times1024$ pixels) using StyleGAN \cite{9107970}. In this approach, facial images are first embedded into StyleGAN’s latent space via a mapping network, after which morphing is performed by computing a weighted combination of the latent codes corresponding to the two subjects. The resulting latent representation is then passed through the synthesis network to generate high-quality morphed images. This methodology was further refined by MIPGAN (Morphing through Identity Prior driven GAN) \cite{zhang2021mipgan}, which incorporates an identity-preserving loss to better retain the biometric characteristics of the contributing subjects. 
More recent deep learning-based morphing approaches generally follow a common pipeline consisting of three main stages: (i) encoding the input images into a latent representation, (ii) interpolating the corresponding latent codes, and (iii) decoding the interpolated representation to generate the finale morphed image. The MorCode approach \cite{pn2024morcode} exploits a Vector Quantized GAN (VQGAN) architecture for the encoding/decoding stage. The recent LADIMO algorithm \cite{LADIMO} exploits the generative power of diffusion models, relying on a Latent Diffusion Model (LDM)-based framework \cite{LDM:CVPR:2022} designed to invert biometric MagFace \cite{meng2021magface} templates and generate morphed images through spherical linear interpolation. 
    
\section{PROPOSED METHOD}\label{sec:method}

\begin{figure*}[t]
    \centering
    \includegraphics[width=\linewidth]{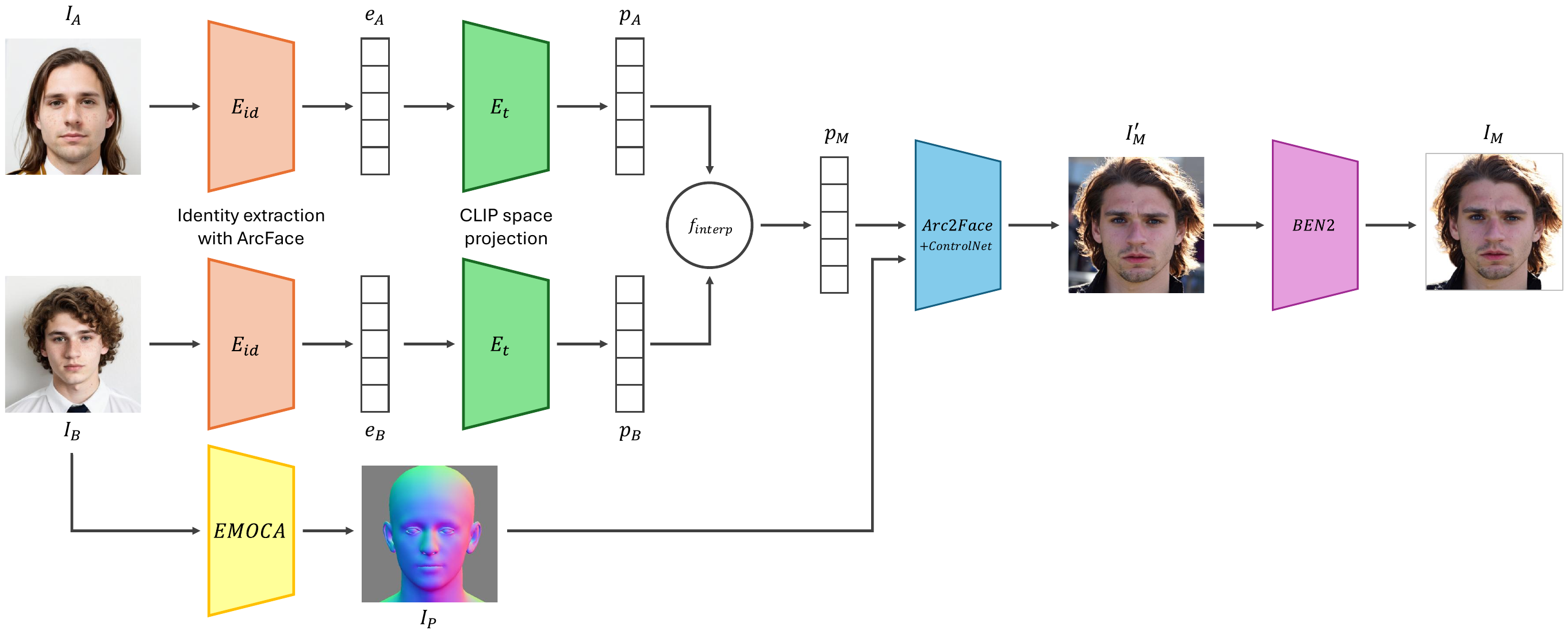}
    \caption{Overview of the proposed method. The input images $I_A$ and $I_B$ are first encoded using the identity encoder $E_id(\cdot)$, producing the identity embeddings $e_A$ and $e_B$.
    In parallel, we extract the pose conditioning image for Arc2Face with EMOCAv2~\cite{danvevcek2022emoca,filntisis2022visual}, obtaining $I_P$. Then, the two compact identity representations $e_A$ and $e_B$ are mapped into the CLIP latent space and interpolated to obtain the latent representation $p_M$, which is then decoded alongside with the conditioning image $I_P$ to generate the morphed image $I_M'$. Finally, the resulting image is post-processed using BEN2 to remove the background, yielding the final morphed image $I_M$.}
    \label{fig:method}
\end{figure*}

Inspired by recent advances in conditional synthetic face generation, we propose a novel face morphing framework based on Arc2Face~\cite{papantoniou2024arc2face}, an identity-conditioned face foundation model capable of synthesizing photorealistic facial images from compact identity representations.
Arc2Face model leverages ArcFace~\cite{deng2019arcface} embeddings as semantic identity priors, thereby enabling controllable and identity-preserving image generation.

Formally, given two input facial images, denoted as $I_A$ and $I_B$, the goal of our method is to generate a morphed image $I_M$ that exhibits a hybrid identity, both perceptually and according to SOTA FRSs.
The proposed pipeline begins by extracting identity features from each input image using a pretrained ArcFace encoder $E_{id}(\cdot)$, producing two normalized 512-dimensional embeddings: $e_A = E_{id}(I_A)$ and $e_B = E_{id}(I_B)$.
These embeddings serve as compact identity representations and constitute the basis for all subsequent processing steps in the proposed pipeline.

Subsequently, the ArcFace embeddings $e_A$ and $e_B$ are projected into the multimodal latent space of CLIP~\cite{radford2021learning} through a modified CLIP text encoder $E_t(\cdot)$. Specifically, each identity embedding is injected into a fixed textual prompt of the form ``photo of a $\langle \text{id} \rangle$ person'', where $\langle \text{id} \rangle$ corresponds to a padded representation of corresponding ArcFace embedding. The encoder processes this prompt and produces a sequence of five token-level embeddings $\left(e_1, e_2, e_3, e_{id}, e_5\right)$, in which the identity information is explicitly encoded as part of the textual conditioning. The resulting CLIP-space representations, $p_A = E_t(e_A)$ and $p_B = E_t(e_B)$, serve as high-level semantic conditioning signals for the image synthesis stage.

To construct the hybrid identity semantic prior, we interpolate between $p_A$ and $p_B$.
Given the multimodal and continuous nature of the CLIP latent space, this interpolation can be performed directly within the CLIP embedding domain, allowing smooth transitions between identities while maintaining fulll compatibility with the downstream generative model.

We define the morphed identity representation as:

\begin{equation}
    p_M = f_{interp}(p_A, p_B, \alpha)
\end{equation}

where $\alpha$ denotes the morphing factor (i.e., the relative contribution of each identity). By default, $\alpha = 0.5$ produces an equal blend of the two identities, while other values can be used to bias the morph toward one of the subject.

Different formulations can be adopted for the interpolation function $f_{\text{interp}}$. In our framework, we consider two widely used strategies: linear interpolation (lerp) \cite{LERP} and spherical linear interpolation (slerp) \cite{SLERP}. Although the CLIP latent space is not explicitly constrained to lie on a hypersphere, CLIP embeddings are L2-normalized and optimized using cosine similarity. Under this assumption, semantic information is predominantly encoded in angular relationships, which makes spherical linear interpolation a reasonable heuristic for identity blending. Linear interpolation, on the other hand, remains a fully valid alternative that does not impose additional geometric assumptions on the structure of the embedding space.

For completeness, the spherical linear interpolation between two vectors $a$ and $b$ is defined as:

\begin{equation}
    \mathrm{slerp}(a, b, t) = \frac{\sin((1-t)\theta)}{\sin{\theta}} a + \frac{\sin(t\theta)}{\sin{\theta}}b
\end{equation}

where $\theta$ denotes the angle between vectors $a$ and $b$.

Regardless of the chosen interpolation function $f_{interp}$, the resulting interpolated CLIP representation $p_M$ is then provided to Arc2Face as the conditioning input, leading to the synthesis of the final morphed image $I_M'$.

Furthermore, to generate morphed images that adhere as much as possible to ISO/ICAO standards~\cite{icao-portrait,ISO-19794-5,ISO-29794-5,ISO-39794-5} with respect to pose, facial expression, and background, we adopt a two-stage control strategy.
First, pose and facial expression are conditioned following the approach proposed in \cite{papantoniou2024arc2face}, by employing a ControlNet~\cite{zhang2023adding} conditioned on a target 3D facial normal map. In our implementation, the condition map ($I_P$) is extracted from the input image $I_B$ using the EMOCAv2 model \cite{danvevcek2022emoca} \cite{filntisis2022visual}, thereby enforcing realistic and identity‑consistent facial geometry in the generated output.
Second, since background appearance cannot be directly controlled in the original Arc2Face model, we apply the BEN2 background removal network~\cite{meyer2025ben}, which estimates an alpha channel isolating the facial region. This allow the original background to be replaced with a uniform white background, ensuring compliance with ISO/ICAO document‑photo requirements.

\section{EXPERIMENTAL EVALUATION}

An extensive experimental evaluation has been conducted to validate the effectiveness of the proposed morphing approach. In particular, the experiments compare the proposed method with several SOTA techniques, both landmark- and deep leaning-based, in terms of attack potential.

\subsection{Datasets}
To provide a comprehensive evaluation of our morphing algorithm, we selected four face datasets, two publicly available and two private, as sources for morph generation. For each dataset, morphed images produced using different existing tools are also available. These datasets therefore serve as benchmarks for comparison, enabling a rigorous assessment of the effectiveness of the proposed algorithm.

\begin{itemize}
    \item FEI \cite{Fei_THOMAZ2010902} and FEI Morph v2 datasets \cite{di2023combining} \cite{FEIMorphDB}: The FEI dataset comprises images of 200 subjects, equally distributed between male and female, with ages mainly ranging from 19 to 40 years. The images exhibit a good variability in terms of appearance, hairstyle, and presence of accessories.
    The FEI Morph v2 dataset was derived from the FEI images using seven SOTA landmark-based morphing algorithms (see Table \ref{tab:MorphingAlgs}) and two morphing factors (0.3 and 0.5), resulting in a total of 14000 morphed images. 
    
    \item ONOT \cite{di2024onot} and MONOT \cite{MONOT} datasets: The ONOT dataset consists of a synthetic collection of high-quality face images designed to comply with ISO/IEC \cite{ISO-39794-5}, and include 254 distinct synthetic subjects.
    The MONOT dataset was generated from ONOT images using six SOTA landmark-based morphing algorithms (see Table \ref{tab:MorphingAlgs}) and two morphing factors (0.3 and 0.5), yielding a total of 15240 morphed images. In addition, for each subject, the MONOT dataset provides ten ”in the wild images” (generated using the Arc2Face model \cite{papantoniou2024arc2face}), to simulate live acquisitions at airport gates.
    Two evaluation protocols are considered for these datasets: (i) a single probe image per subject from the ONOT dataset, and (ii) ten ”in the wild images” probe images from MONOT, used as gate attempts.
    
    \item SOTAMD Digital \cite{SOTAMDDataset} and EINMorph-HQ v2 datasets: The SOTAMD Digital dataset contains high-resolution face images with frontal pose, neutral expression, and controlled illumination, including 300 bona fide images and 1500 gate images collected within the SOTAMD European project \cite{SOTAMDDataset} from 150 subjects of various ethnicities.
    The EINMorph-HQ v2 dataset was derived from the bona fide images of the 150 SOTAMD subjects using six SOTA morphing algorithms (see Table \ref{tab:MorphingAlgs}) and multiple morphing factors, resulting in 9600 morphed images. For morph generation, subject pairs were selected through similarity assessments performed with three commercial SDKs, in order to replicate a realistic and challenging evaluation scenario.

    \item iMARS-MQ and EINMorph-MQ v2 datasets: The iMARS-MQ dataset was collected within the iMARS European project \cite{iMARSPrj} at six different sites, including two international airports (Lisbon and Athens) and four research laboratories. Image acquisition was performed under real border-control conditions using operational Automatic Border Control (ABC) gates. A total of 60 subjects participated in the acquisition, with some individuals captured at multiple locations. Overall, the dataset comprises 205 bona fide images acquired in a controlled setup compliant with for passport-photo requirements, and 612 live gate images captured using real ABC gates.
    The EINMorph-MQ v2 dataset was generated from the bona fide images of the iMARS-MQ subjects using six SOTA morphing algorithms (see Table \ref{tab:MorphingAlgs}) and multiple morphing factors, yielding 4720 morphed images. As in the HQ case, subject pairs were selected based on similarity assessments performed with three commercial SDKs to ensure a realistic and challenging evaluation setting.
\end{itemize}
For all datasets, the same subject pairs defined in the original benchmarks were used to generate morphed images with Arc2Morph, ensuring a direct and fair comparison across all methods. 

\begin{table*}[t]
    \centering
    \small
    \begin{tabular}{|c|c|c|c|c|c|}
    \hline
    \textbf{Algorithm} &
    \textbf{Type} &
    \textbf{FEI Morph v2} &
    \textbf{MONOT} &
    \textbf{EINMorph-HQ v2} &
    \textbf{EINMorph-MQ v2} \\
    \hline
    C01 \cite{FaceMorpher} & L & $\times$ & $\times$ & & \\
    \hline
    C02 \cite{FaceFusion} & L & $\times$ & $\times$ & & \\
    \hline
    C03 \cite{SOTAMDDataset} & L & $\times$ & $\times$ & & \\
    \hline
    C05 \cite{demorphing} & L & $\times$ & $\times$ & & \\
    \hline
    C05-PA97 \cite{demorphing} \cite{RestoreFormer} \tablefootnote{\label{note1}Morphed images generated using the C05 approach \cite{demorphing} were subsequently retouched to remove morphing artifacts by applying a face restoration technique.} & L & & & $\times$ & $\times$ \\
    \hline
    C05-PA98 \cite{demorphing} \cite{GFPGAN} \textsuperscript{\ref{note1}} & L & & & $\times$ & $\times$ \\
    \hline
    C05-PA99 \cite{demorphing} \cite{CodeFormer} \textsuperscript{\ref{note1}} & L & & & $\times$ & $\times$ \\
    \hline
    C08 \tablefootnote{Details about this algorithm cannot be disclosed due to privacy constraints associated with the iMARS European project \cite{iMARSPrj}.}
        & L & $\times$ & & & \\
    \hline
    C15 \cite{INGROUPE} \cite{SURYS} & L & $\times$ & $\times$ & & \\
    \hline
    C16 \cite{C16Paper} & L & $\times$ & $\times$ & & \\
    \hline
    C20 \cite{LADIMO} & D & & & $\times$ & $\times$ \\
    \hline
    C21 \tablefootnote{\label{note2}Details about this algorithm cannot be disclosed due to privacy constraints associated with the EINSTEIN European project \cite{EINSTEINProj}.} & D & & & $\times$ & $\times$ \\
    \hline
    C22 \textsuperscript{\ref{note2}} & D & & & $\times$ & $\times$ \\
    \hline
    \end{tabular}
    \caption{List of the morphing algorithms used to generate the morphed images for the datasets considered in this work. The column \emph{Type} indicates whether the corresponding method is based on facial landmarks (L) or on deep learning techniques (D).}
    \label{tab:MorphingAlgs}
\end{table*}

\subsection{Evaluation metric}
The effectiveness of the proposed morphing algorithm was evaluated by measuring the attack potential of the generated images and comparing it with that obtained from morphed images produced by other SOTA methods. For this purpose, we adopted the Morphing Attack Potential (MAP) metric, introduced in \cite{9794509} and recently incorporated into the ISO/IEC 20059:2025 standard \cite{ISO20059} as the recommended methodology to evaluate the resistance of biometric recognition systems to morphing attacks. MAP quantifies the attack potential (i.e., the probability of successful attacks) of a dataset of morphed images $M$ by jointly considering multiple probe images and multiple FRSs.
MAP is represented as a matrix in which rows correspond to the number of probe images per morph and columns correspond to the number of FRSs involved in the evaluation. Each matrix element $MAP[r,c]$ represents the percentage of morphed images in the dataset $M$ that can be successfully verified with at least $r$ probe images of both contributing subjects by at least $c$ FRSs. 

The two dimensions of the MAP matrix capture key factors for assessing morphing attack potential. The vertical axis represents \textit{robustness}: moving from top to bottom, it reflects the ability of morphed samples to be successfully verified against an increasing number of probe samples. Morphed samples appearing in the lower rows therefore exhibit a robust similarity to both contributing subjects, rather than incidental matches with a single probe.
The horizontal axis represents \textit{generality}: moving from left to right, it evaluates the ability of morphed samples to deceive an increasing number of FRSs. Morphed images in the rightmost columns therefore exhibit a broader and more system-independent attack capability.
The most critical morphed samples are those located in the bottom-right corner of the MAP matrix, as they exhibit both high robustness and high generality. Nevertheless, this region contains only a small number of samples, since relatively few morphs satisfy both conditions simultaneously.

In addition, to facilitate the comparison across different algorithms, the robustness and generality curves are plotted as suggested in the ISO/IEC 20059:2025 standard \cite{ISO20059}. These two curves are computed as the normalized weighted sums of the columns and rows of the MAP matrix, respectively. Specifically, the weight assigned to each row $r$ is $\frac{r}{r_{max}}$ and, analogously, the weight assigned to each column $c$ is $\frac{c}{c_{max}}$ (where $r_{max}$ and $c_{max}$ are the total number of rows and columns in the MAP matrix, respectively).

\subsection{Face Recognition Systems}

For the MAP computation on the FEI Morph v2 and MONOT datasets, and to ensure a fair comparison with the MAP values reported in \cite{MONOT}, the same three Commercial Off-The-Shelf (COTS) FRSs adopted in \cite{MONOT} were employed. These systems were identified as top performers in the Face Recognition Technology Evaluation (FRTE) 1:1 verification benchmark \cite{FRTE}.

Conversely, for the MAP computation on the EINMorph-HQ v2 and EINMorph-MQ v2 datasets, a more comprehensive evaluation protocol was adopted. In this case, six FRSs were considered: the three COTS systems used for FEI Morph v2 and MONOT, together with three SOTA deep learning-based face recognition models (MagFace \cite{meng2021magface}, AdaFace \cite{adaface}, and CurricularFace \cite{CurricularFace}).

The verification thresholds of all FRSs were set to operate at a False Acceptance Rate (FAR) of 0.1\%, which represents the reference operating point for FRSs in electronic Machine Readable Travel Document (eMRTD) applications \cite{FrontexGuidelines}. For the three COTS systems, the thresholds were selected according to vendor recommendations, whereas for the three deep learning models they were calibrated to achieve a FAR of 0.1\% on a subset of the FRGC database \cite{FRGC}.

\subsection{Experimental results}
This section analyzes the results in terms of MAP, evaluated across the different testing datasets.

The results obtained on the FEI Morph v2 dataset are reported in Table~\ref{tab:fei_MAP} where the MAP of the proposed approach is compared with that of other seven SOTA morphing algorithms (see Table~\ref{tab:MorphingAlgs}). 

\begin{table}[t]
    \centering
    \begin{tabular}{c|c|c|c|}
    \cline{2-4}
     & 1 & 2 & 3 \\ \hline
    \multicolumn{1}{|c|}{C01} & 97.9\% & 91.7\% & 74.0\% \\  \hline
    \multicolumn{1}{|c|}{C02} & 99.7\% & 98.9\% & 95.9\% \\  \hline
    \multicolumn{1}{|c|}{C03} & 97.3\% & 89.9\% & 70.2\% \\  \hline
    \multicolumn{1}{|c|}{C05} & 98.4\% & 93.5\% & 78.0\% \\  \hline
    \multicolumn{1}{|c|}{C08} & 98.6\% & 94.4\% & 82.5\% \\  \hline
    \multicolumn{1}{|c|}{C15} & 97.7\% & 91.5\% & 70.9\% \\  \hline
    \multicolumn{1}{|c|}{C16} & 99.3\% & 97.4\% & 88.7\% \\  \hline
    \multicolumn{1}{|c|}{Arc2Morph (ours)} & \textbf{99.9\%} & \textbf{99.7\%} & \textbf{98.7\%} \\  \hline
    \end{tabular}
    \caption{MAP computed on the FEI Morph v2 dataset for the proposed approach and seven competitors. Best results for each column are highlighted in bold. In this case, a single probe image is available, so the MAP of each algorithm is represented by a single row.}
    \label{tab:fei_MAP}
\end{table}

The proposed approach demonstrates outstanding performance in this evaluation, achieving MAP values higher than all other evaluated methods. The results show that 98.7\% of the morphed images generated by the proposed approach deceive all three COTS systems, thereby confirming the effectiveness of the algorithm in preserving identity features.

\begin{table}[]
    \centering
    \begin{tabular}{c|c|c|c|}
    \cline{2-4}
     & 1 & 2 & 3 \\ \hline
     \multicolumn{1}{|c|}{C01} & 94.2\% & \textbf{86.7\%} & 73.5\% \\  \hline
     \multicolumn{1}{|c|}{C02} & 94.1\% & \textbf{86.7\%} & \textbf{74.0\%} \\  \hline
     \multicolumn{1}{|c|}{C03} & 90.9\% & 80.8\% & 63.4\% \\  \hline
     \multicolumn{1}{|c|}{C05} & 91.8\% & 83.0\% & 67.4\% \\  \hline
     \multicolumn{1}{|c|}{C15} & 89.3\% & 78.3\% & 55.8\% \\  \hline
     \multicolumn{1}{|c|}{C16} & 91.8\% & 82.3\% & 61.6\% \\  \hline
     \multicolumn{1}{|c|}{Arc2Morph (ours)} & \textbf{97.6\%} & 86.5\% & 63.6\% \\  \hline
    \end{tabular}
    \caption{MAP computed on the MONOT dataset for the proposed approach and the six competitors reported in \cite{MONOT}, considering a single ONOT image for each subject as probe(the MAP of each algorithm is represented by a single row). Best results for each column are highlighted in bold.}
    \label{tab:monot_MAP_oneprobe}
\end{table}

Table~\ref{tab:monot_MAP_oneprobe} reports the results obtained on the MONOT dataset when a single probe image for each subject (taken from the ONOT dataset) is considered. 
Overall, the observed MAP is slightly lower than that measured on the FEI Morph v2 dataset. This difference is likely due to the characteristics of the datasets: FEI contains real images, whereas ONOT consists of synthetic samples, which tend to exhibit larger intra-subject variations than those typically observed in real-world data, thereby making successful morphing attacks more challenging. 
In this scenario, the proposed approach achieves the highest attack capability when targeting a single FRS, with a success rate exceeding 97\%. Although the MAP decreases as the number of targeted FRSs increases, the proposed method remains comparable to (or outperforms) most competing approaches. Only C01 and C02 show higher generality when all three FRSs are considered.

\begin{table}[]
    \centering
    \begin{tabular}{c|c|c|c|}
    \cline{2-4}
     & 1 & 2 & 3 \\ \hline
    \multicolumn{1}{|c|}{1} & 100.0\% & 100.0\% & 99.9\% \\  \hline
    \multicolumn{1}{|c|}{2} & 100.0\% & 100.0\% & 99.8\% \\  \hline
    \multicolumn{1}{|c|}{3} & 100.0\% & 100.0\% & 99.8\% \\  \hline
    \multicolumn{1}{|c|}{4} & 100.0\% & 100.0\% & 99.6\% \\  \hline
    \multicolumn{1}{|c|}{5} & 100.0\% & 100.0\% & 99.6\% \\  \hline
    \multicolumn{1}{|c|}{6} & 100.0\% & 99.9\% & 99.1\% \\  \hline
    \multicolumn{1}{|c|}{7} & 100.0\% & 99.9\% & 99.1\% \\  \hline
    \multicolumn{1}{|c|}{8} & 100.0\% & 99.8\% & 98.1\% \\  \hline
    \multicolumn{1}{|c|}{9} & 100.0\% & 99.4\% & 95.1\% \\  \hline
    \multicolumn{1}{|c|}{10} & 99.8\% & 96.5\% & 84.3\% \\  \hline
    \end{tabular}
    \caption{MAP computed on the MONOT dataset for the proposed approach, considering ten "in the wild" images for each subject as probe.}
    \label{tab:monot_MAP_tenprobes}
\end{table}

The MAP results in Table~\ref{tab:monot_MAP_tenprobes}, computed using ten "in the wild" probe images for each subject, show an overall very high MAP for the proposed algorithm across all evaluation settings. Even as the number of probe images increases, the attack success rate remains close to saturation for one and two FRSs, and extremely high even when three FRSs are considered, confirming the very high quality of the generated morphed images.
Only under the most restrictive conditions (when a large number of probe images is combined with three FRSs) a performance drop is observed. Nevertheless, even in these cases, the MAP values remain high, indicating that the morphed images preserve a stable similarity with both contributing identities. This consistently high MAP can be partially attributed to the experimental setup: in this scenario, all probe images have been generated using Arc2Face \cite{MONOT}. As a result, the proposed method produces morphed images that are particularly challenging for the FRSs, as they are well aligned with the characteristics of the probe-generation process.

\begin{figure}[htbp]
    \centering
    
    \begin{subfigure}{0.75\linewidth}
        \centering
        \includegraphics[width=\linewidth]{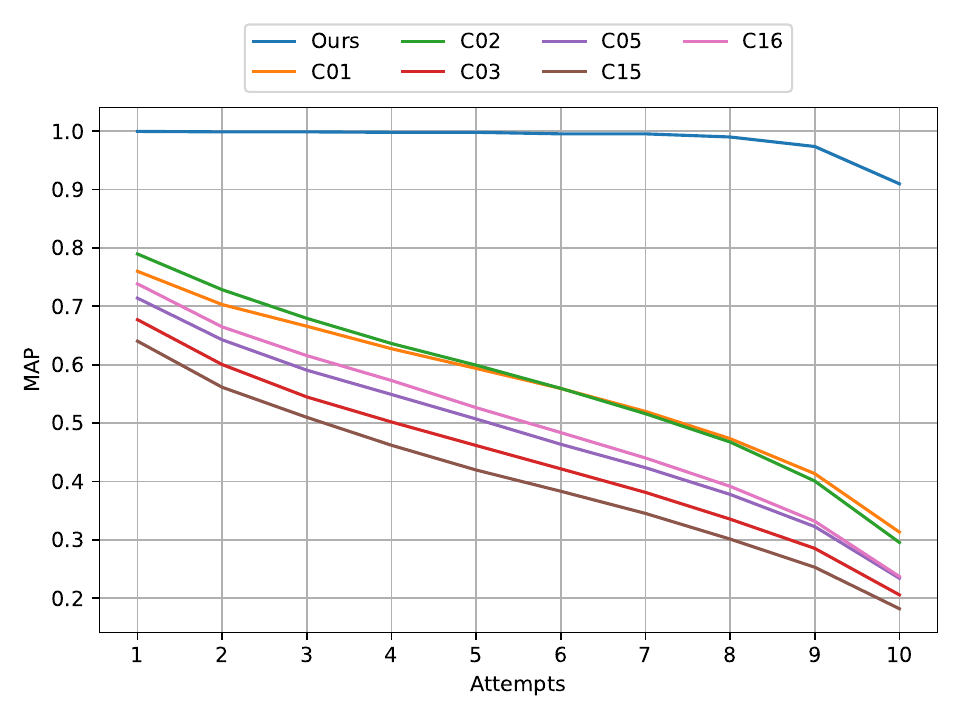}
        \caption{Robustness curve}
        \label{fig:MONOT_vis_robustness}
    \end{subfigure}
    \vfill
    \begin{subfigure}{0.75\linewidth}
        \centering
        \includegraphics[width=\linewidth]{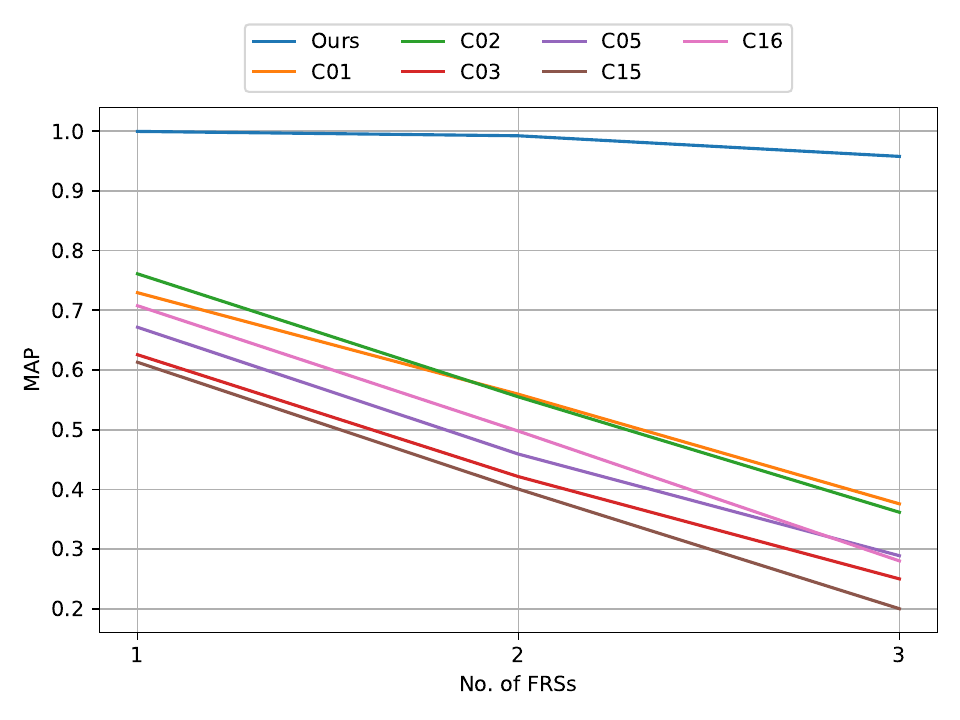}
        \caption{Generality curve}
        \label{fig:MONOT_vis_generality}
    \end{subfigure}
    
    \caption{Visualization of the MAPs computed on the MONOT dataset for the proposed approach and the competitors, considering ten "in the wild" images as probe.}
    \label{fig:MONOT_vis}
\end{figure}

For this testing case, the comparison with the other competitors is provided in concise form in Figure \ref{fig:MONOT_vis}, through the robustness (Figure \ref{fig:MONOT_vis_robustness}) and generality (Figure \ref{fig:MONOT_vis_generality}) curves. The graphs clearly highlight the higher attack potential of the proposed approach with respect to both metrics. As in the previous analysis, the effectiveness is particularly high due to the use of Arc2Face to generate the MONOT "in the wild" probe images.

\begin{figure}[htbp]
    \centering
    
    \begin{subfigure}{0.75\linewidth}
        \centering
        \includegraphics[width=\linewidth]{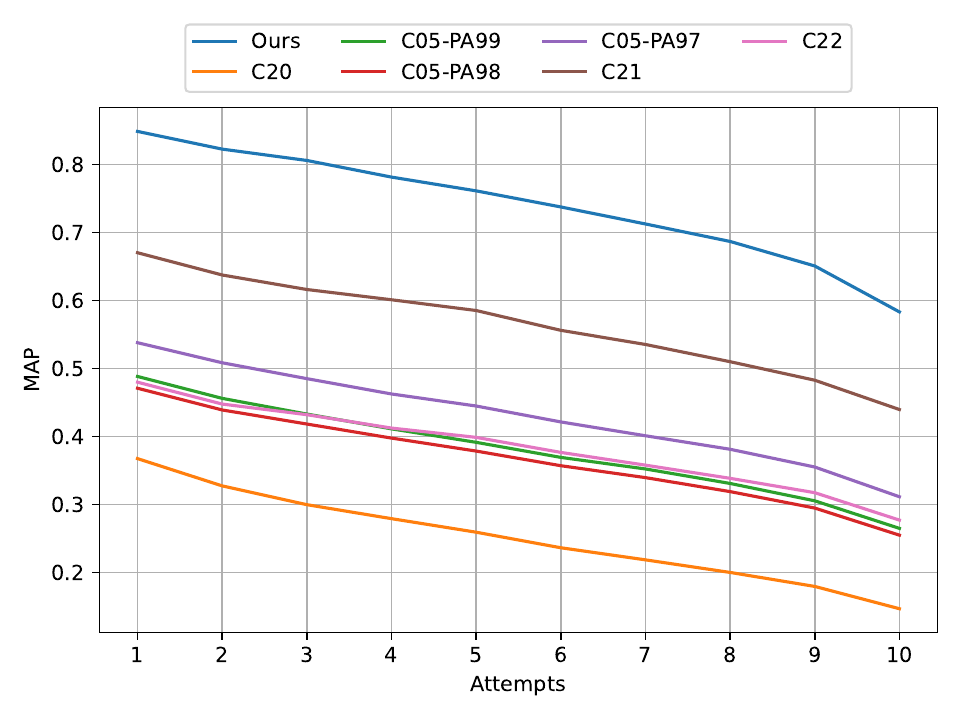}
        \caption{Robustness curve}
        \label{fig:EIN_HQ_robustness}
    \end{subfigure}
    \vfill
    \begin{subfigure}{0.75\linewidth}
        \centering
        \includegraphics[width=\linewidth]{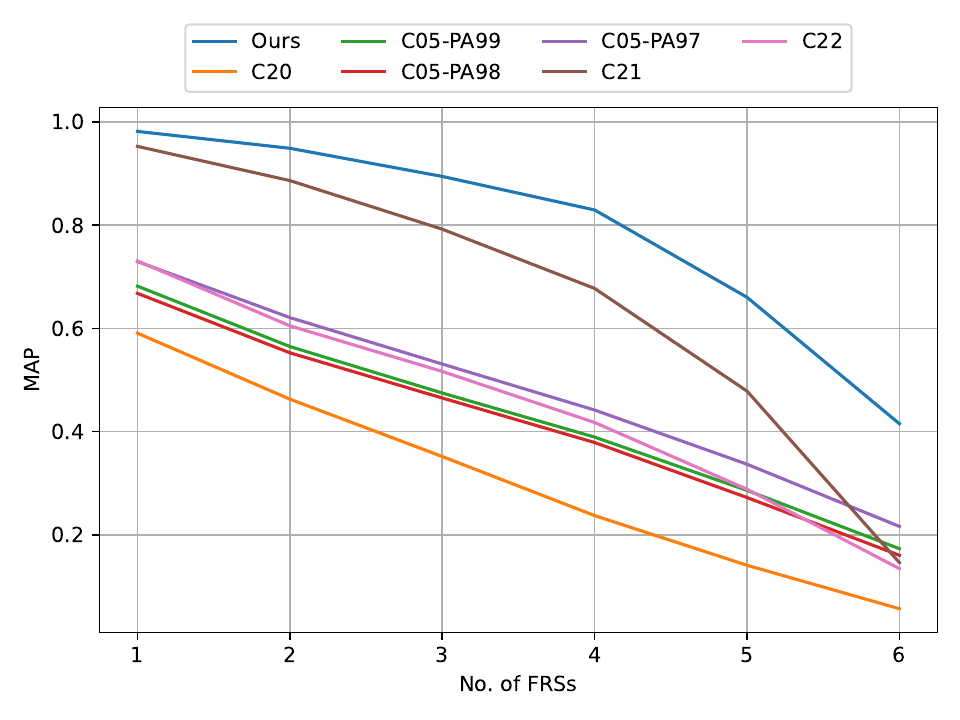}
        \caption{Generality curve}
        \label{fig:EIN_HQ_generality}
    \end{subfigure}
    
    \caption{Visualization of the MAPs computed on the EINMorph-HQ v2 dataset for the proposed approach and the competitors.}
    \label{fig:EIN_HQ_curves}
\end{figure}

\begin{figure}[htbp]
    \centering
    
    \begin{subfigure}{0.75\linewidth}
        \centering
        \includegraphics[width=\linewidth]{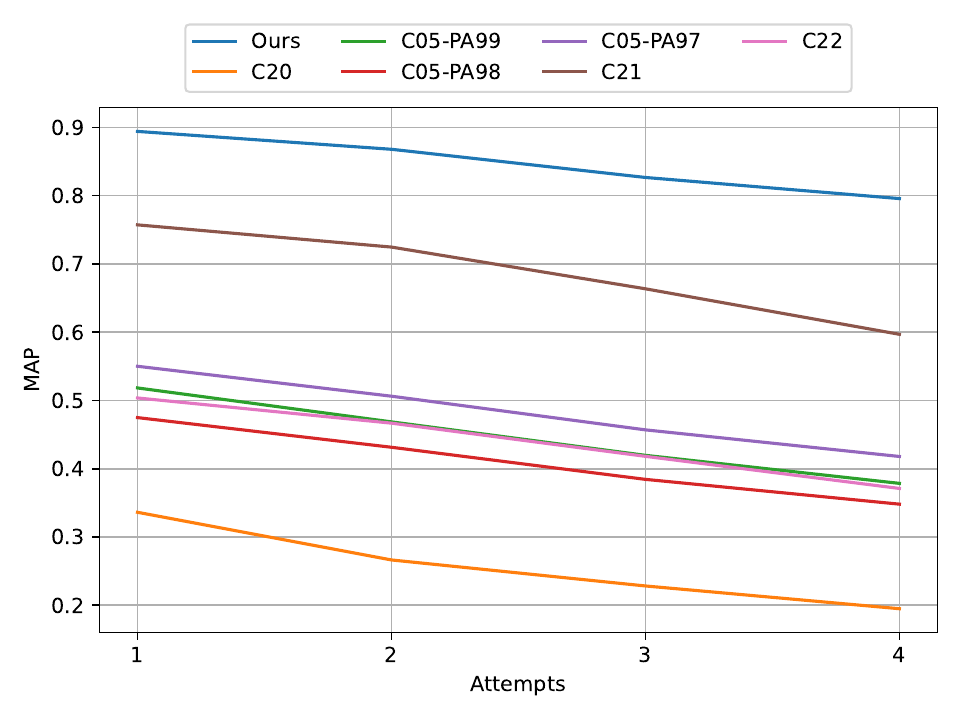}
        \caption{Robustness curve}
        \label{fig:EIN_MQ_robustness}
    \end{subfigure}
    \vfill
    \begin{subfigure}{0.75\linewidth}
        \centering
        \includegraphics[width=\linewidth]{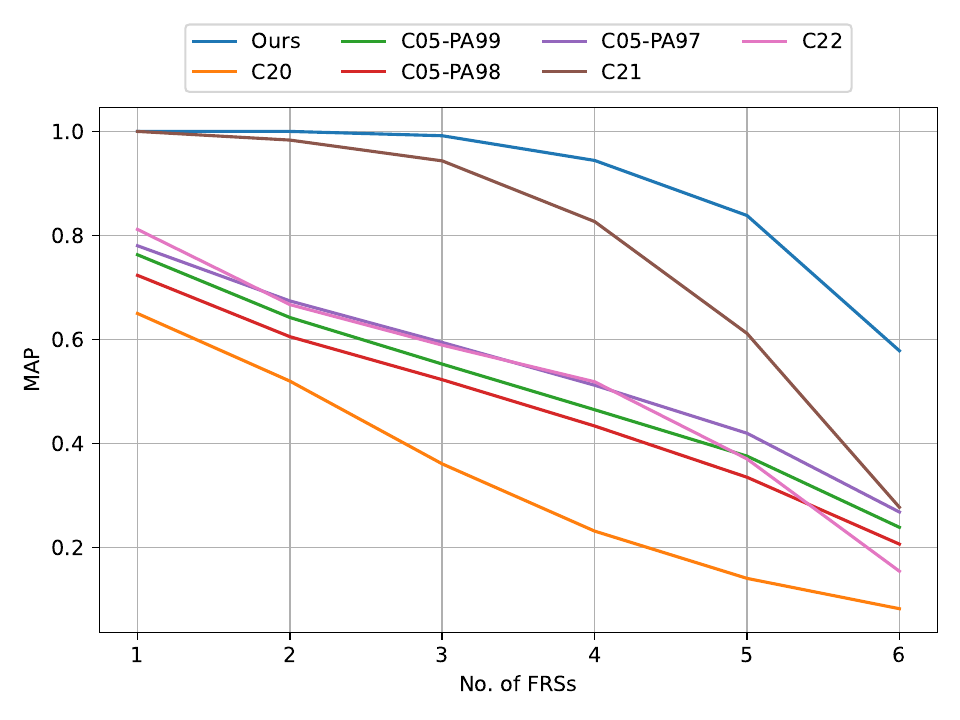}
        \caption{Generality curve}
        \label{fig:EIN_MQ_generality}
    \end{subfigure}
    
    \caption{Visualization of the MAPs computed on the EINMorph-MQ v2 dataset for the proposed approach and the competitors.}
    \label{fig:EIN_MQ_curves}
\end{figure}

A noticeable superiority over the other morphing algorithms, in terms of both robustness and generality, is also confirmed for the EINMorph-HQ v2 (see Figure \ref{fig:EIN_HQ_curves}) and EINMorph-MQ v2 (see Figure \ref{fig:EIN_MQ_curves}) datasets. The curves show that the proposed approach not only surpasses the other deep learning-based algorithms, but also outperforms landmark-based techniques, traditionally considered significantly more challenging for FRSs.

\section{Ablation study: identity interpolation}

\begin{figure}
    \centering
    \begin{subfigure}{0.49\linewidth}
        \centering
        \includegraphics[width=\linewidth]{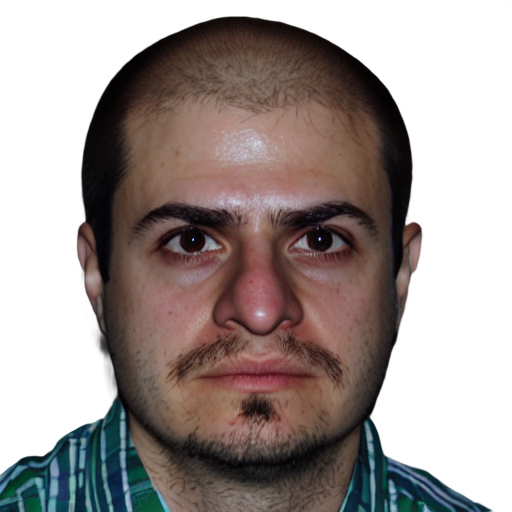}
        \caption{Identity-level lerp}
    \end{subfigure}
    \hfill
    \begin{subfigure}{0.49\linewidth}
        \centering
        \includegraphics[width=\linewidth]{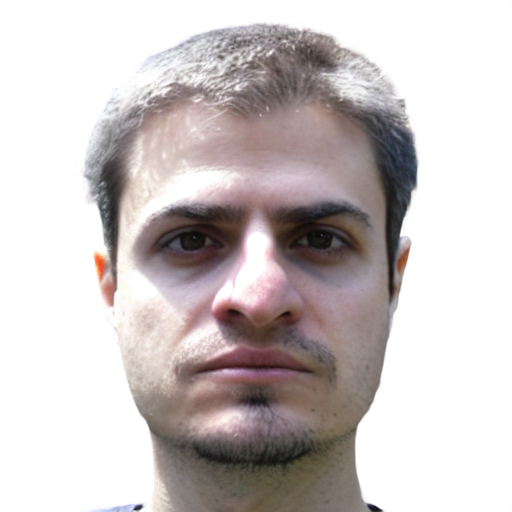}
        \caption{Identity-level slerp}
    \end{subfigure}
    \\
    \vspace{2mm}
    \begin{subfigure}{0.49\linewidth}
        \centering
        \includegraphics[width=\linewidth]{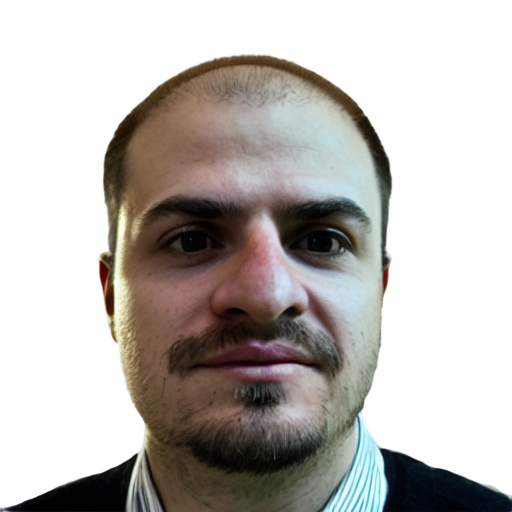}
        \caption{CLIP-level lerp}
    \end{subfigure}
    \begin{subfigure}{0.49\linewidth}
        \centering
        \includegraphics[width=\linewidth]{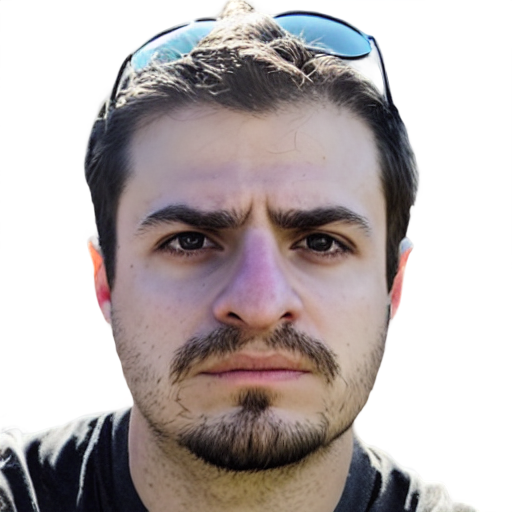}
        \caption{CLIP-level slerp}
    \end{subfigure}
    \caption{A visual example of the morphed images obtained using the different interpolation approaches, given the same image pair as input.}
    \label{fig:interpolation_example}
\end{figure}

As discussed in Section~\ref{sec:method}, several choices can be considered for interpolating the two source identities.
In particular, interpolation can also be applied at the level of identity features, obtaining $e_M = f_{interp}(e_A, e_B, \alpha)$ where $e_A$ and $e_B$ denote the ArcFace identity embeddings of the input images.
The resulting interpolated identity embeddings $e_M$ can then be encoded into the CLIP latent representation $p_M = E_{id}(e_M)$, which is subsequently decoded by Arc2Face to generate the morphed image $I_M'$.
Then interpolation function $f_{interp}$ can be implemented either as linear interpolation (lerp) or spherical linear interpolation (slerp).
Moreover, the interpolation can be performed either in the ArcFace identity space (prior to projection into the CLIP latent space) or directly in the CLIP embedding space. Four combinations are therefore possible, and we compared these alternatives in order to determine the most effective strategy in terms of attack potential. This analysis is carried out on the FEI and ONOT datasets by varying the interpolation method and the embedding space in which it is applied. For each configuration, the MAP is computed.
To facilitate comparison across configurations, MAP values are reported as a single scalar value $MAP_{Avg}$, obtained using the scalarization procedure defined in the ISO/IEC 20059:2025 standard~\cite{ISO20059}.

\begin{table}[t]
    \centering
    \begin{tabular}{cccc}
        \toprule
        \textbf{Dataset} & \textbf{Interp. location} & \textbf{$f_{interp}$} & \textbf{$MAP_{Avg}$} \\
        \midrule
        \multirow{4}{*}{FEI}  & \multirow{2}{*}{Identity} & lerp  & 0.9778 \\
                              &                           & slerp & 0.9679 \\
                              & \multirow{2}{*}{CLIP}     & lerp  & 0.9747 \\
                              &                           & slerp & \textbf{0.9835} \\
        \midrule
        \multirow{4}{*}{MONOT} & \multirow{2}{*}{Identity} & lerp  & 0.8539 \\
                              &                           & slerp & 0.8486 \\
                              & \multirow{2}{*}{CLIP}     & lerp  & 0.8163 \\
                              &                           & slerp & \textbf{0.8858 }\\
        \bottomrule
    \end{tabular}
    \caption{$MAP_{Avg}$ reported on FEI and MONOT datasets for each choice of interpolation location (\textit{i.e.}, identity-level or CLIP latent-level) and interpolation function $f_{interp}$ (\textit{i.e.}, linear interpolation or spherical linear interpolation).}
    \label{tab:interp-map}
\end{table}

As shown in Table~\ref{tab:interp-map}, spherical linear interpolation applied in the CLIP latent space yields the highest overall $MAP_{Avg}$.
We hypothesize that this apparently counterintuitive behavior is due to the higher dimensionality and richer semantic structure of CLIP's latent space, which allow it to capture finer details than the ArcFace embedding space alone. A visual comparison of images generated using the different interpolation strategies is provided in Figure \ref{fig:interpolation_example}.

\section{CONCLUSIONS AND FUTURE WORKS}
This work introduces a new deep learning-based face morphing approach that proved to have an attack potential noticeably higher than that of existing methods, including the landmark-based techniques, which have traditionally been considered as the most challenging for FRSs.

Future research will focus on further improving the morph generation process, with particular attention on the explicit control of additional image characteristics beyond pose, such as illumination, gaze direction, and exposure. Addressing these factors is expected to yield morphed images that are more closely aligned with ISO/ICAO requirements for identity document photographs.

\section*{ETHICAL IMPACT STATEMENT}

This work studies face morphing attacks as a security vulnerability in FRSs for electronic identity documents. The research was conducted without collecting new biometric data and without direct interaction with human participants; therefore, ethical review board oversight was not required under applicable regulations.
All experiments are based exclusively on publicly available and internal-use-only datasets, used in compliance with their intended research purposes and associated consent and privacy constraints. As no new personal data were collected and no individuals were contacted, risks to human subjects are minimal.
The proposed morphing method is developed solely for research and evaluation purposes, with the aim of improving the robustness of morphing attack detection systems. While advanced morphing techniques may pose risks if misused and may reflect limitations in pretrained models, this work does not promote operational deployment of such attacks. Instead, it contributes to strengthening biometric system security by supporting the development of more effective countermeasures.

{\small
\bibliographystyle{ieee}
\bibliography{egbib}
}

\end{document}